\UseRawInputEncoding
\documentclass{article}
\usepackage{spconf,amsmath,graphicx}
\usepackage{cprotect}

\usepackage{color}
\usepackage[colorlinks=true,linkcolor=red,citecolor=green,urlcolor=red]{hyperref}
\usepackage{array}
\usepackage{booktabs}
\usepackage{multirow}
\usepackage{tabularx}
\usepackage[table]{xcolor}
\usepackage{setspace}
\usepackage{caption}
\usepackage{amssymb}
\usepackage{bbding}
\title{HaltingVT: Adaptive token halting transformer for \\ efficient video recognition}
\name{Qian Wu$^*$, Ruoxuan Cui$^*$, Yuke Li$^\dag$, Haoqi Zhu
\thanks{$^*$Equal contributions. $^\dag$Corresponding author.}
}
\address{NetEase Yidun AI Lab, Hangzhou, China}
\email{\{wuqian05, cuiruoxuan, liyuke, zhuhaoqi\}@corp.netease.com}
\begin{document}
\ninept
\maketitle

\begin{abstract}
% Action recognition in videos presents a challenge due to its high computational cost, especially for Joint Space-Time video transformers. Despite their effectiveness, these architectures' excessive number of tokens severely limits their efficiency. In this paper, we propose \textbf{HaltingVT}, which boosts video transformer efficiency from the perspective of dynamic network by a token halting mechanism. HaltingVT applies a data-adaptive token reduction at each transformer layer, achieving a significant reduction in the overall computational cost. Additionally, we introduce the Glimpser module, which quickly removes redundant tokens in shallow transformer layers. This module scans all video patch tokens and removes the most redundant ones, which may even be misleading for video recognition tasks by our observations. To further encourage HaltingVT to focus on the key motion-related information in video, we also design an effective Motion Loss during training. HaltingVT acquires video analysis capabilities and token halting compression strategies simultaneously in a unified training process, without requiring additional training procedures or sub-networks. 
Action recognition in videos poses a challenge due to its high computational cost, especially for Joint Space-Time video transformers 
 (Joint VT). Despite their effectiveness, the excessive number of tokens in such architectures significantly limits their efficiency. In this paper, we propose  \textbf{HaltingVT}, an efficient video transformer adaptively removing redundant video patch tokens, which is primarily composed of a Joint VT and a Glimpser module. Specifically, HaltingVT applies data-adaptive token reduction at each layer, resulting in a significant reduction in the overall computational cost. Besides, the Glimpser module quickly removes redundant tokens in shallow transformer layers, which may even be misleading for video recognition tasks based on our observations. To further encourage HaltingVT to focus on the key motion-related information in videos, we design an effective Motion Loss during training. HaltingVT acquires video analysis capabilities and token halting compression strategies simultaneously in a unified training process, without requiring additional training procedures or sub-networks. 
On the Mini-Kinetics dataset\cite{meng2020ar}, we achieved $\textbf{75.0\%}$ top-1 ACC with $\textbf{24.2}$ GFLOPs, as well as $\textbf{67.2\%}$ top-1 ACC with an extremely low $\textbf{9.9}$ GFLOPs. The code is available at \href{https://github.com/dun-research/HaltingVT}{https://github.com/dun-research/HaltingVT}.

\end{abstract}


\section{Introduction}
\label{sec:intro}

Given the rise in video data usage, concerns over the high computational costs and memory usage during video processing have come to the forefront, prompting many efforts to improve the efficiency of video analysis tasks\cite{luo2019grouped,wu2020adaframepami,gao2020listen,gowda2021smart,lin2022ocsampler,ma2022rethinking}. 

Dynamic network\cite{han2021dynamic,wu2020dynamic2}  is an effective approach for reducing overall computation and improving efficiency by adaptively allocating computational resources based on data characteristics. Specifically, for video analysis networks based on CNN or RNN, various dynamic methods have been explored, including frame sampling \cite{wu2020adaframepami, gowda2021smart, xia2022nsnet, lin2022ocsampler, xia2022tsqnet} and early-exit\cite{ghodrati2021frameexit, wu2020dynamic2}, often with the assistance of additional policy networks \cite{meng2020ar, sun2021videoiq, lin2022ocsampler, 2022_dstep}. When using frame sampling, preserved frames may contain irrelevant information while discarded frames may contain important information, resulting in ``information redundancy \& insufficiency" simultaneously. Besides, frame sampling leads to a lack of spatial-temporal features, which is crucial for processing video tasks both with CNN\cite{gowda2021smart,lin2022ocsampler} and transformer\cite{arnab2021vivit}.

 \begin{figure}[t]
    \centering  
    \includegraphics[width=1.0\linewidth]{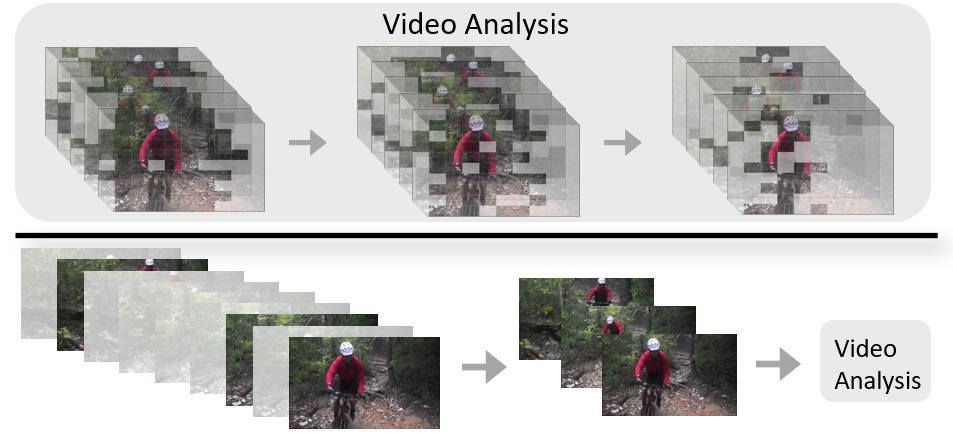}
    \vspace{-2em}
    \cprotect\caption{Illustration of different efficient video analysis strategies. Our HaltingVT (top) reduces overall computation cost by reducing the number of patch tokens, while frame sampling methods (bottom) use a policy network to select key frames, potentially leading to redundant content retention and critical information loss.}
    \vspace{-2em}
    \label{fig:results_plot}
 \end{figure}

\begin{figure*}[t]
    \centering
    \includegraphics[width=0.98\linewidth]{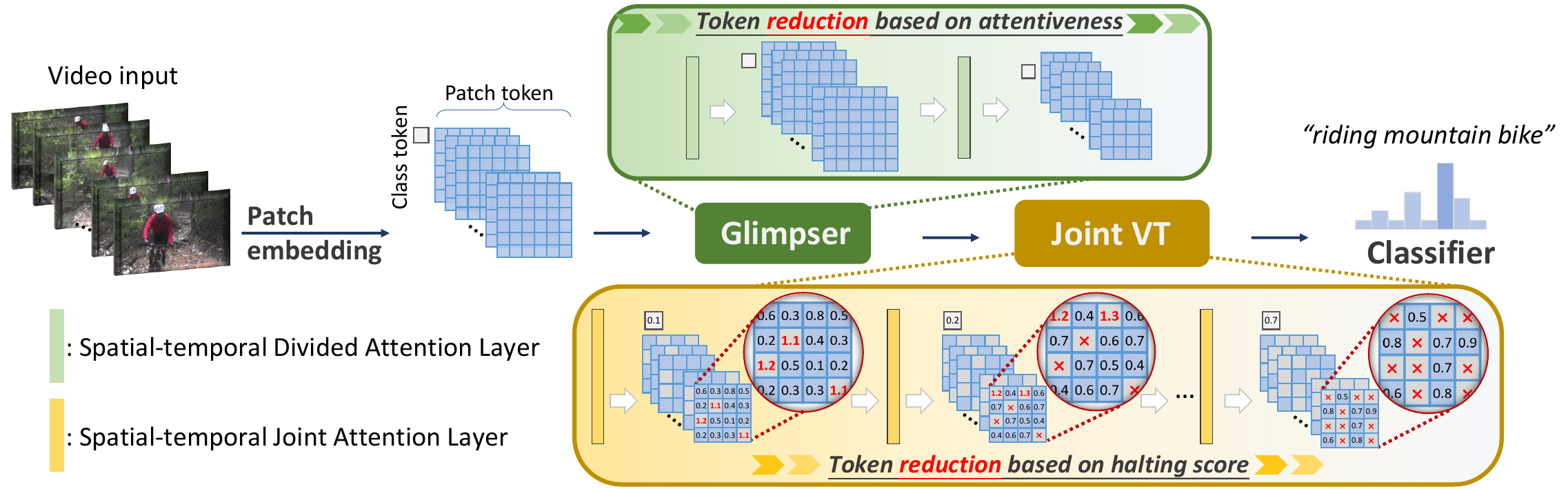}
    \vspace{-2ex}
    \cprotect\caption{HaltingVT takes the Joint Space-Time video transformer with a token halting mechanism as the core model skeleton. We calculate a halting score for each patch token in each transformer layer, and perform token halting when its cumulative halting score exceeds 1 (red cross in the figure). When the class token reaches the final layer or halts, a weighted sum of class tokens based on their halting scores is transmitted to subsequent classifier for the final prediction. The Glimpser removes redundant tokens in the early stage of inference.
    % in the shallow layers.
    % which are the most unrelated to the video task. 
    }
    % \textcolor{red}{add Motion loss in the figure}
    \vspace{-3ex}
    \label{fig:pipeline}
\end{figure*}

Recently, video transformer methods have achieved significant progress\cite{arnab2021vivit,bertasius2021timesformer}. The Joint Space-Time video transformers excel at video analysis tasks, as they can evaluate spatial and temporal attention relationships for all tokens simultaneously. However, video transformers still suffer from high computational costs, with quadratic complexity of token numbers in both space and time $\mathcal{O}(S^2T^2)$. Although some  recent works have optimized the attention calculation in video transformers\cite{liu2022swin,yang2022temporally}, to the best of our knowledge, there is a lack of research on applying dynamic network designs for efficient video transformer architectures. 

Therefore, we propose HaltingVT, a dynamic efficient video transformer architecture based on adaptive token reduction. HaltingVT leverages a data-dependent token halting strategy to reduce the number of tokens layer-by-layer, markedly minimizing computational overhead. To further enhance the computational efficiency and training effectiveness, we also design the Glimpser module and a Motion Loss. These solutions can be applied to other video transformers as well. In summary, our contributions are as follows:

- We propose HaltingVT, a novel adaptive efficient video transformer with token halting, different from previous works focusing on optimizing CNN structures or static video transformer structures.
% which takes a unique approach compared to previous works that focus on optimizing CNN structures or static video transformer structures.

- HaltingVT learns both video processing tasks and token halting decisions in an end-to-end manner, without requiring any additional policy networks or complex reinforcement learning-based training strategies.

- Our comprehensive experiments on well-known benchmarks demonstrate that HaltingVT achieves outstanding performance in both effectiveness and efficiency.

\section{Related works}
\label{sec:format}
\vspace{-2ex}

\noindent
{\bf Efficient video recognition.} In efficient video recognition tasks, a common approach is to select key frames or clips \cite{bhardwaj2019efficient, wu2020adaframepami, korbar2019scsampler, wu2019multi, wu2020dynamic2, xia2022nsnet} from the raw video data for processing.  Many solutions utilize a policy network to make decisions on input data\cite{meng2020ar,ghodrati2021frameexit,lin2022ocsampler,2022_dstep}, network layers\cite{sun2021videoiq}, or other factors\cite{meng2021adafuse}. Achieving appropriate data-adaptive effects often requires complex learning strategies such as reinforcement learning and additional training procedures. Recently, many works have introduced efficient designs for video transformers, such as modifying the attention mechanism\cite{bertasius2021timesformer,liu2022swin,yang2022temporally} or proposing hierarchical models\cite{arnab2021vivit}. Different from the static video architectures with fixed computational graphs and parameters in inference, we propose an efficient video transformer structure from the perspective of dynamic networks by adaptive token halting.

\vspace{0.3em}\noindent
{\bf Halting mechanism.} The halting mechanism\cite{graves2016adaptive} is an approach for dynamic depth networks, originally proposed for RNN models. Recently, this method has been adopted in vision transformers\cite{yin2022avit,ye2023efficient}, where the halting scores of image patch tokens are used for dynamic routing. 
However, applying the halting strategy directly to video frames leads to misalignment of token counts across frames, making it difficult to perform temporal correlation analysis. To address this issue, we design an efficient Joint VT structure with a token halting mechanism for the first time, and propose the Glimpser module to enhance the practical applicability, which leading to further improvements in recognition accuracy and computational efficiency.

% we introduce the Glimpser module to improve the adaptability of the halting mechanism in the Joint Space-Time video transformer. 
% To the best of our knowledge, this study is the first to implement the token halting mechanism in video transformers, resulting in significant improvements in video processing efficiency.

\section{Adaptive video transformer}
\label{sec:pagestyle}

\subsection{Review of video transformers}
\label{3.1_review_video_transformer}
\vspace{-1ex}
% Given a video $V \in \mathcal{R}^{T \times H \times W \times C}$, we get a sequence of 2D patches by 2D convolutions, and then linearly projected them into an embedding space, $x_i \in \mathcal{R}^d$, added with a class token responsible for aggregating global information and final classification. All the tokens are added with positional encodings to inject location information, and then fed to a stack of transformer blocks to compute the spatial and temporal self-attention \textbf{jointly} or \textbf{separately}:

Given a video $V \in \mathcal{R}^{T \times H \times W \times C}$, we obtain a series of patch tokens from a convolutional network, and linearly project them into the token embedding space $X \in \mathcal{R}^{K\times D}$, where ${K}$ represents the total number of patch tokens and $D$ is the embedding dimension.
Along with a class token $\boldsymbol{x}_c$ to capture global information and facilitate final classification, all tokens are injected with positional encodings and fed into subsequent transformer blocks to compute spatial and temporal self-attention \textbf{jointly} or \textbf{dividedly}:

% \vspace{1em}\noindent
\vspace{-3ex}
\begin{equation}
\label{eq_attention}
\mathrm{Attention}(\mathbf{Q}, \mathbf{K}, \mathbf{V})=\mathrm{Softmax}(\mathbf{QK}^\top/\sqrt{C}) \mathbf{V} ,
\end{equation}
\vspace{-3ex}

\noindent
where $\mathbf{Q}$, $\mathbf{K}$, and $\mathbf{V}$ represent the query, key, and value embeddings based of ${X}$, respectively.

\vspace{0.3em}\noindent
{\bf Joint Space-Time Video Transformer(Joint VT)} is a straightforward video extension of ViT\cite{dosovitskiy2020image},
% which models pairwise interactions between all tokens in \ref{eq_attention}, 
computing the spatial and temporal self-attention between all tokens jointly. Joint VT exhibits superior performance in analyzing shift motion features and long-range token associations, but with increased computation budget. 
% \vspace{0.3em}

\vspace{0.3em}\noindent
{\bf Divided Space-Time Video Transformer (Divided VT)} is an factorized video transformer architecture\cite{bertasius2021timesformer}, where temporal and spatial attention are applied separately in consecutive steps. Divided VT is lightweight but may overlook certain motion information in the video data that involves spatiotemporal shifts.

\vspace{-2ex}
\subsection{Adaptive token halting mechanism}
\label{3.2halting_mechanism}
\vspace{-1ex}

After a 2D embedding network, we adopt an $L$-layer Joint VT,  and calculate a halting score\cite{graves2016adaptive, yin2022avit} for every token at every layer via Eq. (\ref{eq_halting_score}). The halting scores are used to guide token halting operations, thereby reducing the number of tokens layer by layer.

% and $\sigma(u)=\frac1{1+\exp^{-u}}$ is the logistic sigmoid function:

% \vspace{-1ex}\noindent
\vspace{-1ex}
\begin{equation}
\label{eq_halting_score}
h_k^l=\sigma(\gamma\cdot x_{k,0}^l+\beta) 
\end{equation}
\vspace{-2ex}

\noindent
Here  $x_{k,0}^l$ is the first dimension of $\boldsymbol{x}_k^l$ , which indicates a patch token $\boldsymbol{x}_k$ at layer $l$. The halting score is calculated by logistic sigmoid function $\sigma(u)$, in which $\gamma$ and $\beta$ are used to adjust the magnitude of score values, thereby influencing the degree of token reduction and the performance of video recognition.

% \noindent

We ``halt'' the token at the $N_k$-th layer once its cumulative halting score surpasses $1-\epsilon$, signifying this token's computation ends here and it will not be propagated to subsequent layers. A small positive constant $\epsilon$ is used to allow token halting after the first layer:

% \vspace{0.3em}\noindent
\vspace{-2ex}
\begin{equation}
\label{eq_halt_condition}
N_k=\operatorname*{argmin}_{n\leq L}\sum_{l=1}^nh_k^l\geq 1-\epsilon .
\end{equation}
\vspace{-1ex}
% \noindent

\noindent
When the computation reaches the final layer, we define $h^L_{1:K} = 1$ to enforce stopping for all tokens. We compute a ``reminder" for each token via Eq. (\ref{r_k}), which refers to its ``escape threshold" at the halting layer $N_K$. To encourage the model to learn a reasonable token halting mechanism, a ponder loss is designed as Eq. (\ref{ponderloss}).

% \vspace{0em}\noindent
\vspace{0em}\noindent
\begin{equation}
\label{r_k}
r_k=1-\sum_{l=1}^{N_k-1}h_k^l 
\end{equation}
\vspace{-0.2em}

\vspace{-0.2em}\noindent
\begin{equation}
\label{ponderloss}
    \mathcal{L}_{\text{ponder}}:=\frac{1}{K}\sum_{k=1}^{K}(N_{k}+r_{k}) 
\end{equation}
\vspace{-1ex}

\noindent
As for the video task loss $\mathcal{L}_\text{task}$, we first get $\boldsymbol{x}_o$, a weighted sum of class token $\boldsymbol{x}_c$ based on its halting scores from all preceding layers. Then, $\boldsymbol{x}_o$ is passed to the subsequent video recognition classifier $\mathcal{C}$:

\vspace{-3ex}
\begin{equation}
\label{classification}
    \mathcal{L}_\text{task}=L_{CE}(\mathcal{C}(\boldsymbol{x}_o)),\text{ where }\boldsymbol{x}_o=\sum_{l=1}^{N_c-1}h_c^l\boldsymbol{x}_c^l + r_c\boldsymbol{x}_c^{N_c} .
\end{equation}

% \vspace{-2ex}
\subsection{Glimpser module}
\label{3.3glimpser}
\vspace{-1ex}
% The computational cost of Joint VT has quadratic complexity of token numbers in both space and time $\mathcal{O}(S^2T^2)$. 
The token halting mechanism in Sec. \ref{3.2halting_mechanism} effectively reduces the overall number of tokens, but potentially redundant information in shallow layers can be removed more efficiently as a preliminary filtering step. To address this issue, we propose ``Glimpser'', an upstream module for HaltingVT, which rapidly scans all tokens and removes the most redundant ones. Glimpser consists of two Divided Space-Time transformer layers, and we calculate each patch token's self-attention to the class token $\boldsymbol{x}_c$ in the second layer via: 
% \vspace{1em}\noindent
\begin{equation}
\label{attentiveness}
\boldsymbol{x}_{\text {c}}= \mathrm{Softmax}(\mathbf{\boldsymbol{q}_{\text {c}}\boldsymbol{K}}^\top/\sqrt{C}) \mathbf{V}=\boldsymbol{w} \cdot 
\boldsymbol{V} ,
\end{equation}

\noindent
where the class token $\boldsymbol{x}_c$  is a linear combination of the existing patch value embeddings $\boldsymbol{V}=[\boldsymbol{v}_1,\boldsymbol{v}_2,\ldots,\boldsymbol{v}_K]$, and $\boldsymbol{w}$ represents the ``attentiveness" of existing patch tokens with respect to  $\boldsymbol{x}_c$ . We adopt the top-K principle by retaining only the patch tokens with the highest attentiveness values, proportional to a ratio $R$. This simple operation does not introduce any additional computational overhead, but significantly reduces the computational burden of the subsequent Joint VT module by $R^2\times$. 

% It is worth noting that the experimental results in \ref{ablation} demonstrate that our Glimpser module can accurately identifies the redundant tokens, which could even be detrimental to the video analysis task. This not only significantly reduces computational costs, but also enhances overall task performance.

% It is noteworthy that the experimental results in Sec. \ref{ablation} indicate that our Glimpser module can accurately identify the redundant tokens in shallow networks, which could even be harmful to the video task. The Glimpser module not only significantly reduces the computational cost but also further enhances video recognition performance.

It is noteworthy that the visualization in Sec. \ref{sec:visualization} indicates that our Glimpser module can accurately identify the redundant tokens in shallow layers, which could even be harmful to the video task as demonstrated by the experimental results in Sec. \ref{ablation}. Thus the Glimpser module not only significantly reduces the computational cost but also further enhances video recognition performance.

% Using the top-k principle, we remove the patch tokens with the lowest attentiveness value, which are the least important to the video task. The remaining patch tokens, along with the class token, are then fed into the subsequent Joint encoder.

\vspace{-2ex}
\subsection{Motion Loss}
\label{3.4motionloss}
\vspace{-1ex}
To amplify the inherent advantage of Joint VT in extracting motion features with temporal and spatial offsets, we propose a straightforward and effective training objective called ``Motion Loss" for the training process. We first expand the original single training sample to a ``training sample pair" consisting of two piece of data:

(i) $V$: the real video data, labeled as ``positive";

(ii) $V_{fake}$: a stack of duplicated frames sampled from $V$, labeled as ``negative".

% \noindent
$V$ and $V_{fake}$ have identical formats and sizes, differing only in that $V$ 
contains genuine shift motion cues and scene features in video, while  $V_{fake}$ just includes context features from images. The Motion Loss function utilizes a Cross-Entropy loss as shown in Eq. (\ref{motion_loss}) to guide the model to distinguish between authentic and fake video data. This ensures that HaltingVT focuses on regions where crucial actions take place, rather than solely relying on the semantic information from images.
\vspace{0em}
\begin{equation}
\label{motion_loss}
    \mathcal{L}_\text{motion}=L_{CE}(y, y')
\end{equation}
% Taking into account both the video task loss, responsible for video analysis, and the ponder loss, guiding token halting, the overall loss calculation equation for HaltingVT is the weighted sum presented in (\ref{overall_loss}), with weight parameters $\alpha_\text{p}$ and $\alpha_\text{m}$.
Considering the video recognition loss $\mathcal{L}_\text{task}$ and token halting loss $\mathcal{L}_\text{ponder}$ previously mentioned,  the overall loss calculation equation for HaltingVT is the weighted sum presented in Eq. (\ref{overall_loss}), with weight parameters $\alpha_\text{p}$ and $\alpha_\text{m}$.

\vspace{-1ex}
\begin{equation}
\label{overall_loss}    \mathcal{L}_\text{overall}=\mathcal{L}_\text{task}+\alpha_\text{p}{ \mathcal{L}_\text{ponder}}+\alpha_\text{m}{ \mathcal{L}_\text{motion}}
\end{equation}

% \vspace{-1ex}
\section{EXPERIMENTS}
\label{sec:typestyle}
% \vspace{-1ex}

\vspace{-1ex}
\subsection{Experimental Setup}
\vspace{-1ex}

% \vspace{0.3em}
\noindent
\textbf{Datasets.} We evaluate the proposed method on two datasets: Mini-Kinetics\cite{meng2020ar} and ActivityNet-v1.3\cite{caba2015activitynet}. 
% Since the full Kinetics\cite{kay2017kinetics} dataset, which is the most commonly used dataset for video tasks, is quite large, we use its subset, Mini-Kinetics\cite{meng2020ar}, following the most methods for efficient video inference \cite{meng2020ar, korbar2019scsampler, wang2021adafocus}. 
Since the full Kinetics\cite{kay2017kinetics} is quite large, we use its subset, Mini-Kinetics\cite{meng2020ar}, following most methods for efficient video inference \cite{meng2020ar, korbar2019scsampler, wang2021adafocus}. 
Mini-Kinetics comprises 200 action categories and contains 121K videos for training and 10K videos for testing, with each video lasting 6-10 seconds.
ActivityNet-v1.3 is a widely used video dataset for multi-label action recognition tasks. It includes 200 human activity categories, with 10,024 training videos and 4,926 validation videos. Due to the long average video duration of 117 seconds, ActivityNet is more challenging, particularly for the methods without frame sampling.
% Since the videos are long, with an average duration of 117 seconds and may have multiple labels, the dataset is more challenging.

\noindent\textbf{Evaluation metrics.} %We use top-1 accuracy (Acc) for multi-class classification on Mini-Kinetics and mean average precision(mAP) for multi-label classification on ActivityNet for validation, respectively. To measure the computational cost, we use giga floating-point operation (GFLOPs), which is hardware-independent. For all experiments, we report video-level GFLOPs because recognition methods use varying numbers of frames per video. 
We validate performance on Mini-Kinetics using top-1 accuracy (ACC) and multi-label classification performance on ActivityNet using mean average precision (mAP). To measure computational cost, we use giga floating-point operations (GFLOPs), which are hardware-independent. For all experiments, we report GFLOPs per video for comparison since some previous methods use different numbers of frames per video.

\noindent\textbf{Implementation details.} We uniformly sample 8 and 10 frames per video for Mini-Kinetics and ActivityNet, respectively. The input frames are random rescaled and cropped to $224 \times 224$ in training, while are rescaled to $256$ and center cropped to $224 \times 224$ during inference. Besides, we adopted RandAug\cite{cubuk2020randaugment} for training. 
%所有实验中，我们均采用vit-small结果作为backbone，head num为6。对动态token选择，我们使用了两阶段的训练策略，首先在divid+joint backbone上静态训练，再在此基础上使用早停机制动态训练。第一阶段中我们总共训练80个epoch，设置learning rate为3e-5，第二阶段learning rate为1e-5，共训练30个epoch，两阶段均采用余弦退火衰减机制。
% For all of our dynamic learning model, we use the DeiT-T\cite{touvron2021deit} as the backbone, and pretrain the base video model from publicly available checkpoint pretrained on ImageNet\cite{deng2009imagenet} and initialize our HaltingVT models with the base video model.
In all experiments, we pretrain a Base Video Model from the publicly available Deit-S\footnote{\url{https://github.com/facebookresearch/deit}}, and then initialize our HaltingVT models with the Base Video Model for dynamic training.
%, which has similar computational complexity to ResNet50\cite{he2016resnet}, and adopt a two-stage training strategy following\cite{graves2016adaptive}. 
% Specifically, we first train the model with vit-small weights pretrained on ImageNet\cite{deng2009imagenet} without halting mechanism and then initialize the halting model with the pretrained weights for the second stage training. 
Unless otherwise specified, we set $\gamma$ to 10, $\beta$ to 10, $\alpha_p$ to $5 \times 10^{-4}$ and $\alpha_m$ to 0.01, and adopt Adam optimizer and cosine annealing decay with initial learning rate as $1\times10^{-5}$. Additional details are available in our code.
% Total epoch are set as $20$ and $160$ for Mini-Kinetics and ActivityNet, respectively. Adam optimizer\cite{kingma2014adam} with the weight decay as $1e-5$ are used.
For clarity, we use ``\textit{HaltingVT$^{\beta}$($R$)}'' to denote a HaltingVT model employing the Glimpser ratio of $R$ and halting mechanism with parameter $\beta$. 
Additionally, $\beta$ can regulate the model's sensitivity to redundant tokens, as described in Eq. (\ref{eq_halting_score}).
% a smaller $\beta$  make the model more sensitive. 
% Thus we use different $\beta$ value for different model seeking for their best performance.

% % \vspace{-2ex}
% \begin{table}[h]
% \small
% \setstretch{0.5}
% \begin{tabular}{lcc}
% % {>
% % {\left\arraybackslash}m{0.5\linewidth} >{\centering\arraybackslash}m{0.15\linewidth} >{\centering\arraybackslash}m{0.15\linewidth}}
% \toprule
% \multicolumn{1}{c}{\textbf{Approach}} & \textbf{Top1(\%)} & \textbf{GFLOPs/V} \\ 
% \midrule
% % \rowcolor[HTML]{C0C0C0} 
% % \quad Joint-ST ViT & 76.3 & 56.5 \\
% \quad HaltingVT$^{10}$ & 71.5 & 31.8 \\ 
% \midrule
% \quad Ours$^{10}$~+~Glimpser(0.3) & 71.0 & 14.9 \\
% \quad Ours$^{10}$~+~Glimpser(0.5)~ & 72.4 & 21.1 \\
% \quad Ours$^{10}$~+~Glimpser(0.7) & 73.3 & 26.3 \\ 
% \midrule
% % \rowcolor[HTML]{C0C0C0} 
% \quad Ours$^{10}$~+~Glimpser(0.3)~+~ML & 71.6 & 14.1 \\
% \quad Ours$^{10}$~+~Glimpser(0.5)~+~ML & 72.5 & 20.5 \\
% \rowcolor[HTML]{C0C0C0} 
% \quad Ours$^{10}$~+~Glimpser(0.7)~+~ML & 74.0 & 26.1 \\ 
% \bottomrule
% \end{tabular}
% \vspace{-2ex}
% \caption{Ablation results on Mini-Kinetics. The superscripted number represents the value of $\beta$, ``Glimpser(*)'' denotes the ratio $R$ of Glimpser, ``ML" denotes the model trained with the Motion Loss.}
% \vspace{-3ex}
% \label{table_ablation}
% \end{table}

\vspace{-2ex}
\begin{table}[h]
\centering
\small
\setstretch{0.5}
\begin{tabular}{cc|cc}
% >{\columncolor[HTML]{FFFFFF}}c 
% >{\columncolor[HTML]{FFFFFF}}c 
% >{\columncolor[HTML]{FFFFFF}}c 
% >{\columncolor[HTML]{FFFFFF}}c }
\toprule
{ \textbf{Glimpser}} & { \textbf{Motion Loss}} & { \textbf{Top1(\%)}} & { \textbf{GFLOPs/V}} \\
\midrule
{ \XSolidBrush} & { \XSolidBrush} & { 71.5} & { 31.8} \\
\midrule
{ $R = 0.3$} & { \XSolidBrush} & { 71.0} & { 14.9} \\
{ $R = 0.5$} & { \XSolidBrush} & { 72.4} & { 21.1} \\
{ $R = 0.7$} & { \XSolidBrush} & { 73.3} & { 26.3} \\
\midrule
{ $R = 0.3$} & { \Checkmark} & { 71.6} & { 14.1} \\
{ $R = 0.5$} & { \Checkmark} & { 72.5} & { 20.5} \\
\rowcolor[HTML]{C0C0C0} 
{ $R = 0.7$} & { \Checkmark} & { 74.0} & { 26.1} \\
\bottomrule
\end{tabular}
\vspace{-2ex}
\caption{Ablation results on Mini-Kinetics. We compare the HaltingVT models with (\Checkmark) or without (\XSolidBrush) Glimpser / Motion Loss.}
\vspace{-4ex}
\label{table_ablation}
\end{table}

\vspace{-1ex}
\subsection{Ablation Studies}
\label{ablation}
\vspace{-1ex}

%在这个部分，我们通过消融实验来验证我们所提出的HaltingVT、Glimpser和Motion Loss在提升视频分类效果与效率上是否有效，实验结果在table1中展示。
% In this section, we conduct ablation experiments to validate the effectiveness of our proposed HaltingVT, Glimpser, and Motion Loss in improving the accuracy and efficiency of video classification. The experimental results are presented in Table \ref{table_ablation}.

% \noindent\textbf{Efficiency of the HaltingVT.} %我们首先在

\noindent\textbf{Effectiveness of the Glimpser.} %我们在Tab1中对比了有无Glimpser的我们的模型的结果，并且探究了Glimpser中不同keep rate对结果的影响。有Glimpser的模型的推理效率均显著优于没有Glimpser的模型，当keep_rate=0.3时，推理效率可提升2.25x，而Acc仅收到非常微弱的影响。随着keep_rate增加到0.7，acc可以提升2.3%，但推理效率仍然有1.2x的提升。这符合我们前面的推测，在推理的前期去除大量明显冗余的token，不仅有利于提升推理效率，同时也能促进模型将注意力集中于有效信息上，从而达到更好的分类效果。
% We compared the results of our model with and without Glimpser in Table \ref{table_ablation} and investigated the impact of different keep rates in Glimpser on the results. The inference efficiency of the model with Glimpser was significantly better than that without Glimpser, with a 2.25x improvement when $\mathcal{R}=0.3$, while the Acc was only slightly affected. As $\mathcal{R}$ increased to $0.7$, the Acc improved by $2.3\%$, but the inference efficiency still had a $1.2\times$ improvement. This is consistent with our previous speculation that removing a large number of obvious redundant tokens in the early stage of inference is not only conducive to improving inference efficiency but also promotes the model to focus attention on effective information, thus achieving better classification performance.
We conducted experiments on HaltingVT with/without Glimpser, and analyzed the impact of different keep ratio $R$ in Glimpser, as shown in Table \ref{table_ablation}. 
The results indicate that using Glimpser significantly improved the inference efficiency of the model, achieving a \textbf{2.13$\times$} acceleration at $R=0.3$, with minimal effect on accuracy.
%The results indicate that using Glimpser results in significantly improved inference efficiency for the model, achieving a \textbf{2.13$\times$} improvement at $R=0.3$, with minimal impact on accuracy.
% The results indicate that the model with Glimpser achieved significantly better inference efficiency than the model without Glimpser, with an improvement of \textbf{2.13$\times$} when $R=0.3$, while the accuracy was only slightly affected. 
As $R$ increased to $0.7$, the accuracy improved by \textbf{1.8\%} while keeping a $1.2 \times$ efficiency improvement. 
% The results support our hypothesis that removing obvious redundant tokens in the early stage of inference can improve inference efficiency and help the model focus on effective information, resulting in better classification performance.
% This result supports our observation that redundant tokens can even be detrimental to action recognition tasks, which may be due to that they can blur the model's focus, particularly under the global attention mechanism of transformer. 
These findings supports our observation that redundant tokens could be detrimental to action recognition tasks, possibly because they may obscure the model's focus, especially under the global attention mechanism of transformers. 
% Therefore, halting these tokens can improve recognition performance as well as inference efficiency.

\noindent\textbf{Effectiveness of the Motion Loss.} 
As shown in Table \ref{table_ablation}, using Motion Loss during training improved accuracy by  $0.1\%-0.7\%$ while reducing computational cost by up to $\textbf{5.4\%}$, which may be due to Motion Loss leading the models to capture motion information more effectively, requiring fewer tokens to be inspected.
% Moreover, simultaneously incorporating Glimpser and Motion Loss in HaltingVT can improve accuracy by $2.5\%$ while also reduce computational cost by $18.9\%$.
% The results presented in Table \ref{table_ablation} demonstrate that incorporating Motion Loss during training leads to a significant improvement in accuracy, with gains ranging from $0.1\%-0.7\%$, while also reducing computational cost by up to $\textbf{5.4\%}$ ($14.9\rightarrow \textbf{14.1}$). This improvement may be attributed to Motion Loss leading the models to capture motion information more effectively, requiring fewer tokens to be inspected. 
Furthermore, the simultaneous incorporation of Glimpser and Motion Loss in HaltingVT yields an even greater improvement in accuracy of $\textbf{2.5\%}$, while also reducing computational cost by $\textbf{17.9\%}$ ($31.8\rightarrow \textbf{26.1}$).

% 此外，在table2中可以看到，当beta更大时，motions with motion loss 能够有更显著的效果提升。事实上，根据eq.(2）可知，beta能够调节模型对冗余token的敏感度，beta越大，越多的token会被认定为冗余并早停，从而使推理效率更高，这以前的方法中也被验证过。

% Additionally, a larger $\beta$ leads to a greater improvement in accuracy with Motion Loss, as decipted in Table \ref{table_sota_mink}. In fact, $\beta$ can regulate the model's sensitivity to redundant tokens, as described in Eq. (\ref{eq_halting_score}). The model with a larger $\beta$  may identify more tokens as redundant and halt them, thereby enhancing the inference efficiency.
%as verified in previous works\cite{yin2022avit}.

% \vspace{-5ex}
\begin{figure}[t]
\centering
\includegraphics[width=0.95\linewidth]{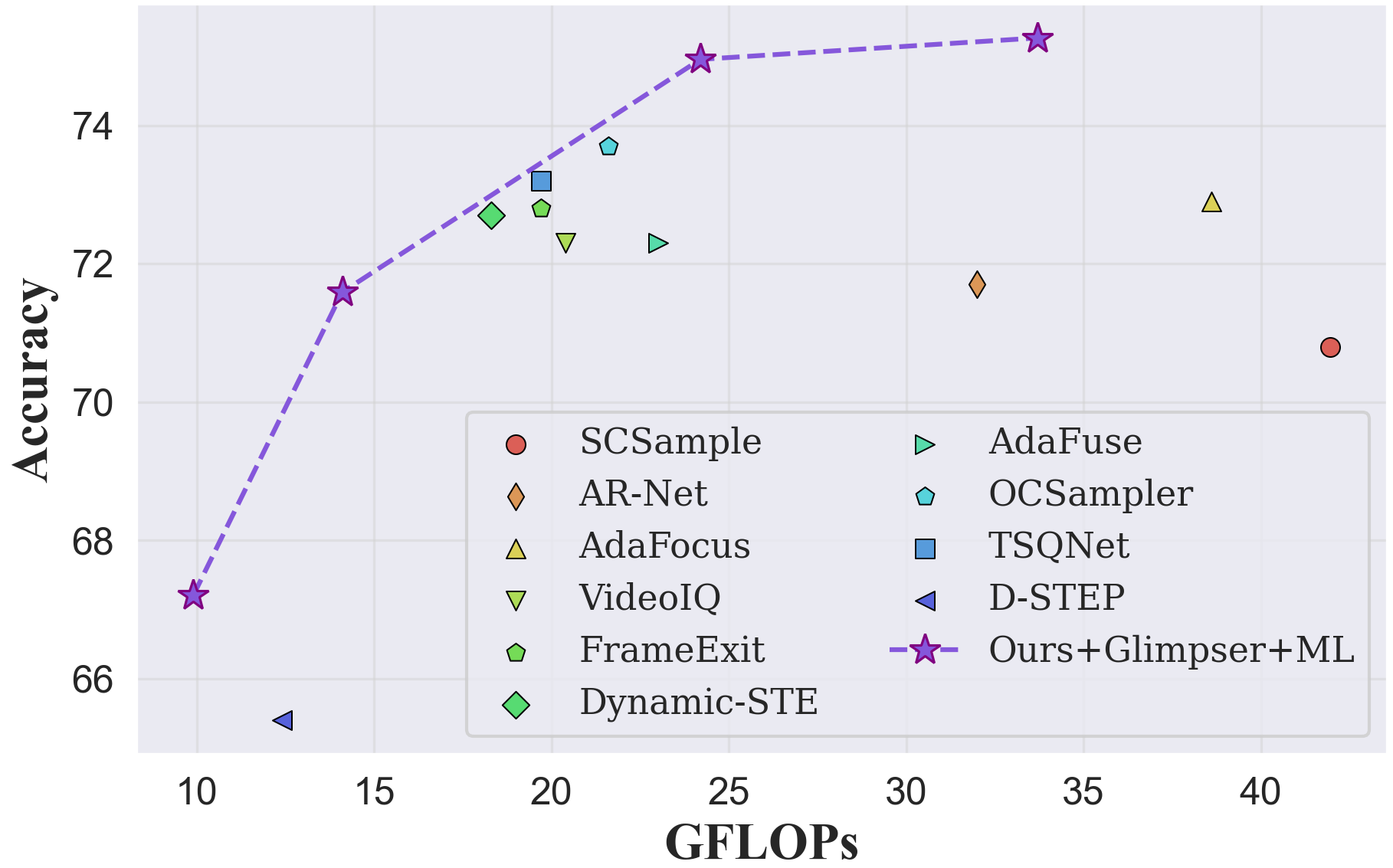}
    \vspace{-2.5ex}
    \caption{
    % Video Recognition performance comparison on accuracy and GFLOPs. 
    Comparison of video recognition performance in terms of accuracy and GFLOPs on Mini-Kinetics.
    HaltingVT achieves the best trade-off between efficiency and effectiveness against SOTAs.}
    \vspace{-4ex}
    \label{fig:res_plot}
\end{figure}

% \vspace{-2ex}
\subsection{Comparison with the state-of-the-art methods.}
\label{sec:comparison_sota}
% \vspace{-1ex}
\noindent\textbf{Mini-Kinetics.} 
% To the best of our knowledge, we are the first to apply data-adaptive token halting/reduction mechanism on video transformer using dynamic training, which is a meaningful attempt. 
% We compare our method with other SOTA efficient video analysis methods, including those based on frame sampling\cite{meng2020ar,wu2019adaframe, meng2021adafuse} or early-exit\cite{ghodrati2021frameexit}.
% As shown in the last two rows of Table \ref{table_sota_mink}, our method achieves the best classification performance with a lead of up to $\textbf{1.6\%}$ ($73.7\% \rightarrow \textbf{75.3\%}$), while maintaining the GFLOPs in a competitive range. 
% Furthermore, our HaltingVT maintains relatively stable performance in scenarios where low computational requirements are critical, for example, ``\textit{Ours$^{5}$~+~Glimpser(0.3)~+~ML}'' outperforms D-STEP by $1.8\%$, while further reducing computational cost by $\textbf{20.1\%}$ ($12.4$G $\rightarrow \textbf{9.9G}$).
To the best of our knowledge, we are the first to apply data-adaptive token halting/reduction mechanism on video transformer using dynamic training, which is a meaningful attempt. 
We compare our method with other state-of-the-art (SOTA) efficient video analysis methods, including those based on frame sampling\cite{meng2020ar,wu2020adaframepami, meng2021adafuse} or early-exit\cite{ghodrati2021frameexit}.
As shown in the last two rows of Table \ref{table_sota_mink}, our method achieves the best classification with a lead of up to $\textbf{1.6\%}$ ($73.7\% \rightarrow \textbf{75.3\%}$) over the SOTA\cite{lin2022ocsampler} while maintaining the GFLOPs in a competitive range.
% our method achieves the best classification performance with a lead of up to $\textbf{1.6\%}$ ($73.7\% \rightarrow \textbf{75.3\%}$), while maintaining the GFLOPs in a competitive range. 
Furthermore, HaltingVT maintains relatively stable performance in scenarios where low computational requirements are critical, for example, ``\textit{HaltingVT$^{5}$(0.3)}'' outperforms D-STEP by $1.8\%$ ($65.4\% \rightarrow \textbf{67.2\%}$), while further reducing computational cost by $\textbf{20.2\%}$ ($12.4$G $\rightarrow \textbf{9.9G}$). The experimental results suggest that HaltingVT achieves outstanding performance across a wide range of computational costs.

\vspace{-1ex}
\begin{table}[h]
\small
\setstretch{0.7}
\captionsetup{font={stretch=0.5}} % 设置表格标题行距为正常行距的 0.8 倍
\centering
\begin{tabular}{>{\centering\arraybackslash}m{0.28\linewidth} >{\centering\arraybackslash}m{0.15\linewidth} >{\centering\arraybackslash}m{0.15\linewidth} >{\centering\arraybackslash}m{0.2\linewidth}} 
\toprule
% \multirow{2}{*}{\textbf{Approach}} & \multirow{2}{*}{\textbf{Publication}} & \multicolumn{2}{c}{\textbf{Mini-Kineticss}} \\
% & & \textbf{Top1(\%)} & \textbf{GFLOPs/V} \\
% \cmidrule(lr){3-4}
\textbf{Approach} & \textbf{Publication} & \textbf{Top1(\%)} & \textbf{GFLOPs/V} \\
\midrule
LiteEval\cite{wu2019liteeval} & NeurIPS'19 & 61.0 & 99.0 \\
SCSample\cite{korbar2019scsampler} & ICCV'19 & 70.8 & 41.9 \\
AR-Net\cite{meng2020ar} & ECCV'20 & 71.7 & 32.0 \\
AdaFocus\cite{wang2021adafocus} & ICCV'21 & 72.9 & 38.6 \\
VideoIQ\cite{sun2021videoiq} & ICCV'21 & 72.3 & 20.4 \\
FrameExit\cite{ghodrati2021frameexit} & CVPR'21 & 72.8 & 19.7 \\
Dynamic-STE\cite{kim2021dynamicste} & ICCV'21 & 72.7 & 18.3 \\
AdaFuse\cite{meng2021adafuse} & ICLR'21 & 72.3 & 23.0 \\
OCSampler\cite{lin2022ocsampler} & CVPR'22 & 73.7 & 21.6 \\
TSQNet\cite{xia2022tsqnet} & ECCV'22 & 73.2 & 19.7 \\
D-STEP\cite{2022_dstep} & BMVC'22 & 65.4 & 12.4 \\ 
\midrule
% \multicolumn{2}{l}{ \ Ours~+~Glimpser(0.3)} & 71.01 & 14.9 \\
% \multicolumn{2}{l}{ \ Ours~+~Glimpser(0.5)} & 73.02 & 20.6 \\
% \multicolumn{2}{l}{ \ Ours~+~Glimpser(0.7)} & 73.27 & 26.3 \\
\multicolumn{2}{l}{ \ \ HaltingVT$^{5}$ (0.3)} & 67.2 & \textbf{9.9} \\
\multicolumn{2}{l}{ \ \ HaltingVT$^{10}$ (0.3)} & 71.6 & 14.1 \\
\rowcolor[HTML]{C0C0C0} 
\multicolumn{2}{l}{ \ \ HaltingVT$^{20}$ (0.5)}  & 75.0 & 24.2 \\
\multicolumn{2}{l}{ \ \ HaltingVT$^{20}$ (0.7)}  & 75.3 & 33.7 \\
\bottomrule
\end{tabular}
\vspace{-0.8em}
\caption{
Comparation with SOTAs on Mini-Kinetics. The bold number indicates the minimum computational cost.
% Comparation with SOTAs on Mini-Kinetics. 
% HaltingVT$^{\beta}$ ($R$) indicates the $\beta$ and Glimpser Ratio of the model.
% The superscripted number of ``HaltingVT'' represents the value of $\beta$, ``$(\bullet)$'' denotes the ratio $R$ of Glimpser. The bold number indicates the minimum computational cost. 
% ``*'' on ``Ours'' means $\beta=20$, otherwise $\beta=10$.
}
\vspace{-1ex}
\label{table_sota_mink}
\end{table}

\begin{table}[t]
\small
\setstretch{0.7}
\captionsetup{font={stretch=0.5}}
    \centering
    \begin{tabular}{>{\centering\arraybackslash}m{0.25\linewidth} >{\centering\arraybackslash}m{0.2\linewidth} >{\centering\arraybackslash}m{0.15\linewidth} >{\centering\arraybackslash}m{0.2\linewidth}} 
    \toprule
    % \multirow{2}{*}{\textbf{Approach}} & \multirow{2}{*}{\textbf{Publication}} & \multicolumn{2}{c}{\textbf{ActivityNet}} \\
    % \cmidrule(lr){3-4}
    % & & \textbf{Top1(\%)} & \textbf{GFLOPs/V} \\
    \textbf{Approach} & \textbf{Publication} 
    & \textbf{mAP(\%)} & \textbf{GFLOPs/V} \\
    \midrule
        LiteEval\cite{wu2019liteeval} & NeurIPS'19 & 72.7 & 95.1 \\
        SCSample\cite{korbar2019scsampler} & ICCV'19 & 72.9 & 42.0 \\
        AR-Net\cite{meng2020ar} & ECCV'20 & 73.8 & 33.5 \\
        ListenToLook\cite{gao2020listen} & CVPR'20 & 72.3 & 81.4 \\
        AdaFrame\cite{wu2020adaframepami} & T-PAMI'20 & 71.5 & 79.0 \\
        AdaFuse\cite{meng2021adafuse} & ICLR'21 & 73.1 & 61.4 \\
        VideoIQ\cite{sun2021videoiq} & ICCV'21 & 74.8 & 28.1 \\
        AdaFocus\cite{wang2021adafocus} & ICCV'21 & 75.0 & 26.6 \\
        OCSampler\cite{lin2022ocsampler} & CVPR'22 & 77.2 & 25.8 \\ 
        \midrule
        % \multicolumn{2}{l}{\quad Ours$^{10}$~+~Glimpser(0.3)} & 70.3 & 19.1 \\
        % \multicolumn{2}{l}{\quad Ours$^{10}$~+~Glimpser(0.5)} & 73.6 & 27.8 \\
        % % \rowcolor[HTML]{C0C0C0} 
        % \multicolumn{2}{l}{\quad Ours$^{10}$~+~Glimpser(0.7)} & 74.2 & 36.0 \\
        % \midrule
        \multicolumn{2}{l}{\quad HaltingVT$^{10}$ (0.3)} & 72.0 & \textbf{19.0} \\
        \multicolumn{2}{l}{\quad HaltingVT$^{10}$ (0.5)} & 73.4 & 26.5 \\
        % \multicolumn{2}{l}{\quad Ours$^{20}$~+~Glimpser(0.7)~+~ML} & 74.5 & 42.8 \\
        \rowcolor[HTML]{C0C0C0}
        \multicolumn{2}{l}{\quad HaltingVT$^{20}$ (0.5)} & 75.4 & 32.5 \\
        \bottomrule
    \end{tabular}
    \vspace{-2ex}
    \caption{Comparation with SOTAs on ActivityNet. The bold number indicates the minimum computational cost.}
    \vspace{-4ex}
    \label{table_sota_anet}
\end{table}

\noindent\textbf{ActivityNet.} 
Table \ref{table_sota_anet} presents a comparison of our method with SOTA approaches on ActivityNet.
% We also compare our method with SOTA approaches on ActivityNet, as shown in Table \ref{table_sota_anet}. 
Due to the long duration of the ActivityNet dataset, the methods with frame sampling have a natural advantage over other methods, such as \cite{lin2022ocsampler, wang2021adafocus, meng2020ar, sun2021videoiq} in Table \ref{table_sota_anet} speed up video
inference by sampling a very few frames, but generally rely on an additional policy
network.
However, our models achieve competitive performance at low computational cost even without an explicit key frame selection mechanism, especially, our ``\textit{HaltingVT$^{10}$(0.3)}'' model maintains relatively stable performance at the extremely low GFLOPs as $\textbf{19.0}$. 
In fact, our token halting approach is compatible with existing keyframe sampling techniques, and incorporating these techniques could potentially enhance the performance of our models. Importantly, our primary focus is on exploring the innovative dynamic video token halting mechanism, thus we simply utilized uniform frame sampling for clarity.
% Furthermore, we plan to investigate how to achieve end-to-end adaptive frame sampling and token selection through dynamic training in the future.
% It is worth noting that our primary focus is on exploring the novel dynamic video token halting mechanism, and thus we utilized uniform frame sampling for simplicity. Meanwhile, some existing keyframe sampling techniques are compatible with our approach, and we believe they can further improve the performance of our models. 
% Moreover, our future work will include exploring a dynamic inference strategy that incorporates data-adaptive frame sampling while token halting mechanism is in effect.

% \vspace{-2ex}
\subsection{Visualization}
\label{sec:visualization}
\begin{figure}[h]
    \centering
    \includegraphics[width=0.95\linewidth]{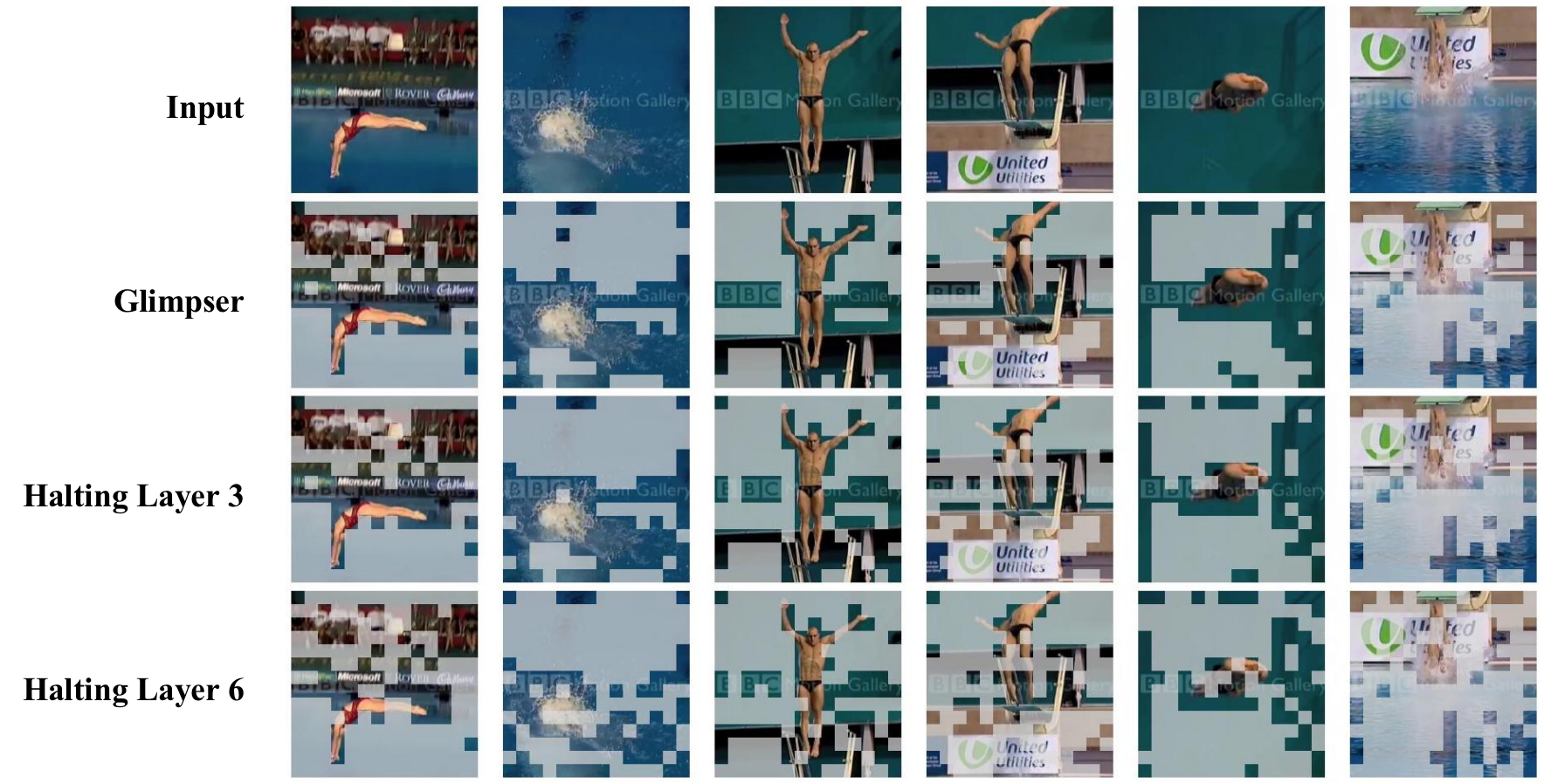}
    \vspace{-2ex}
    \caption{Visualization for token halting.  
    %每行帧序列依次为input，Glimpser阶段的token selction结果，已经HaltingVT中第三和第六层的token halting结果。
    % Each row represents the input, token selection result from the Glimpser, and token halting result from the third and sixth layers of HaltingVT, respectively.
    Each row represents the token halting status of different layers. The patches with lower brightness indicate the tokens that have been halted at that layer.}
    \vspace{-3ex}
    \label{fig:visualization}
\end{figure}

\vspace{-1ex}
We further visualize the token halting process during inference. Fig. \ref{fig:visualization} illustrates a video sequence of a \textit{Diving} sample.
During the Glimpser stage, numerous obvious redundant patches are initially removed. At this stage, token selection is based on fixed proportions, but relatively conservative, ensuring that areas containing task-relevant information are preserved. 
% In the subsequent layers, redundant tokens halt layer-by-layer and also various between different frames, for example, more tokens are redundant and halted in the second and fourth frames than others.
In subsequent layers, redundant tokens are progressively halted layer-by-layer and also adaptively vary between different frames. For example, more tokens are redundant and halted in the second and fourth frames than others.

\vspace{-0.8em}
% \vspace{-1ex}

% \vspace{-1ex}
\section{CONCLUSION}
\label{sec:copyright}
% \vspace{-1ex}

Despite significant development in video transformers, excessive redundant tokens result in high computational costs and reduced efficiency. To address this issue, we propose HaltingVT, an adaptive efficient video transformer architecture with a token halting mechanism. Unlike prior methods that enhance the efficiency of video CNN models via frame extraction or early-exit, we optimize the joint video transformer from the perspective of dynamic networks with adaptive token halting mechanism. Additionally, we propose the Glimpser module and Motion Loss, achieving a balance between effectiveness and efficiency in an end-to-end training process. These approaches can also be extented to other video transformers. Looking ahead, we plan to apply HaltingVT to additional video analysis tasks that demand high efficiency.

% Although video transformers have undergone significant development, excessive redundant tokens result in high computational costs and reduced efficiency. To address this issue, we propose HaltingVT, an adaptive efficient video transformer architecture with token halting mechanism. Unlike prior methods that enhance the efficiency of video CNN models via frame extraction or early-exit, we optimize the joint video transformer from the perspective of dynamic networks by token halting mechanism. By introducing a data-dependent token halting strategy, HaltingVT gradually reduces the number of patch tokens in each video transformer layer, boosting overall computational efficiency. Additionally, we propose the Glimpser module and Motion Loss, further achieving a balance between effectiveness and efficiency in an end-to-end training process. These approaches can also be applied to other video transformers. Looking ahead, we plan to apply HaltingVT to additional video analysis tasks that demand high efficiency.

% Below is an example of how to insert images. Delete the ``\vspace'' line,
% uncomment the preceding line ``\centerline...'' and replace ``imageX.ps''
% with a suitable PostScript file name.
% -------------------------------------------------------------------------

\vfill
\pagebreak

\bibliographystyle{IEEEbib}
\begin{spacing}{0.95}
    \small
    \bibliography{strings}
\end{spacing}

% {
% \small
% \bibliography{strings}
% }

\end{document}